\pdfoutput=1

\documentclass[11pt]{article}

\usepackage{authblk}
\usepackage[]{emnlp2021}

\usepackage{times}
\usepackage{latexsym}
\usepackage{amsmath}
\usepackage{soul}
\usepackage{graphicx}
\usepackage{booktabs}
\usepackage{multirow}
\usepackage{amssymb}
\usepackage{pifont}
\newcommand{\cmark}{\ding{51}}%
\newcommand{\xmark}{\ding{55}}%
\usepackage{array}

\usepackage[T1]{fontenc}

\usepackage[utf8]{inputenc}

\usepackage{microtype}

\newcommand{\hs}[1]{\textcolor{blue}{Huan: #1}}

\newcommand{\add}[1]{\textcolor{red}{#1}}

\newcommand{\revise}[1]{#1}

\newcommand{\nop}[1]{}
\newcommand{\ours}[0]{\text{ReasonBERT}}
\newcommand\std[1]{\ensuremath{\scriptstyle \pm #1}}

\title{\ours: Pre-trained to Reason with Distant Supervision}

\author[1]{Xiang Deng\thanks{Corresponding authors.}}
\author[1]{Yu Su}
\author[2]{Alyssa Lees}
\author[2]{You Wu}
\author[2]{Cong Yu}
\author[1]{Huan Sun\footnote[1]}

\affil[1]{%
The Ohio State University, Columbus, OH}
\affil[ ]{\texttt {\normalsize \{deng.595,su.809,sun.397\}@osu.edu}}
\affil[2]{%
Google Research, New York, NY}
\affil[ ]{\texttt {\normalsize \{alyssalees,wuyou,congyu\}@google.com}}

\begin{document}
\maketitle
\begin{abstract}

We present \ours, a pre-training method that augments language models with the ability to reason over long-range relations and multiple, possibly hybrid, contexts. Unlike existing  pre-training methods that only harvest learning signals from local contexts of naturally occurring texts, we propose a generalized notion of distant supervision to automatically connect multiple pieces of text and tables to create pre-training examples that require long-range reasoning. Different types of reasoning are simulated, including intersecting multiple pieces of evidence, bridging from one piece of evidence to another, and detecting unanswerable cases.
We conduct a comprehensive evaluation on a variety of extractive question answering datasets ranging from single-hop to multi-hop and from text-only to table-only to hybrid that require various reasoning capabilities and show that
\ours\ achieves remarkable improvement over an array of strong baselines. Few-shot experiments further demonstrate that our pre-training method substantially improves sample efficiency.\footnote{Our code and pre-trained models are available at \url{https://github.com/sunlab-osu/ReasonBERT}.}

\end{abstract}

\section{Introduction}

\begin{figure*}[ht]
    \centering
    \includegraphics[width=0.99\linewidth]{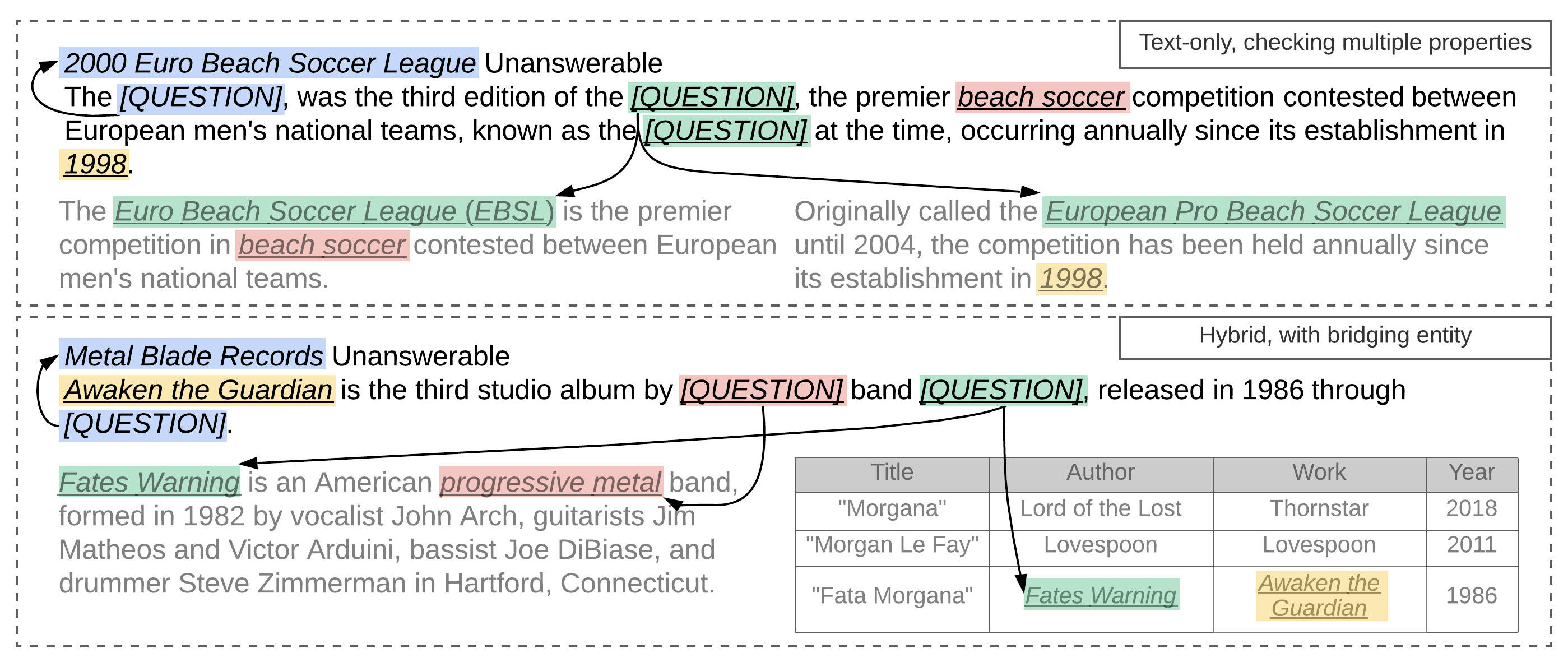}
    \caption{Examples of our pre-training data acquired via distant supervision, which covers a wide range of topics with both textual and tabular evidence. For each query sentence (in black), we first select two pairs of entities (underlined) to find two pieces\nop{selections} of evidence (in grey) via distant supervision. We then randomly mask one entity from each selected pair and aim to recover it by reasoning over the evidence. Note that the two selected pairs may share a common entity; in case this entity is masked, we can mimic different types of multi-hop reasoning, e.g., intersection (Ex. 1) and bridging (Ex. 2). To simulate unanswerable cases, we additionally mask one entity (in blue) that does not exist in the evidence. Figure best viewed in color.
}
    \vspace{-1em}
    \label{fig:example}
\end{figure*}

Recent advances in pre-trained language models (LMs) have remarkably transformed the landscape of natural language processing. Pre-trained \nop{to reconstruct naturally occurring utterances sampled from massive text corpora }with self-supervised objectives such as autoregressive language modeling \cite{Radford2018ImprovingLU,Radford2019LanguageMA,NEURIPSgpt3} and masked language modeling (MLM) \cite{devlin-etal-2019-bert, Liu2019RoBERTaAR, joshi-etal-2020-spanbert}, LMs encode a great deal of knowledge about language and significantly boost model performance on a wide range of downstream tasks \cite{liu-etal-2019-multi,NEURIPS2019_superglue,wang-etal-2018-glue} ranging from spell checking \cite{awasthi-etal-2019-parallel} to sentiment analysis \cite{xu-etal-2019-bert} and semantic parsing \cite{rongali2020don}, just to name a few.  

Existing self-supervised objectives for LM pre-training primarily focus on consecutive, naturally occurring text. For example, MLM enables LMs to correctly predict the missing word ``\textit{daughters}'' in the sentence ``\textit{Obama has two \_\_ , Malia and Sasha.}'' based on the local context and the knowledge stored in the parameters. However, many tasks require reasoning  beyond local contexts: multi-hop question answering (QA) \cite{yang-etal-2018-hotpotqa, welbl-etal-2018-constructing} and  fact verification \cite{jiang-etal-2020-hover} require reasoning over multiple pieces of evidence, hybrid QA \cite{chen-etal-2020-hybridqa} requires simultaneously reasoning over unstructured text and structured tables, and dialogue systems require reasoning over the whole dialogue history to accurately understand the current user utterance \cite{SMDataflow2020}. 

To address this limitation in existing LM pre-training, we propose \ours, a pre-training method to augment LMs for explicitly reasoning over long-range relations and multiple contexts. \nop{Unlike existing pre-training objectives that predict individual masked tokens or spans within a contiguous paragraph of text, }\ours\ pairs a query sentence with multiple relevant pieces of evidence drawn from possibly different places and defines a new LM pre-training objective, \textit{span reasoning}, to recover entity spans that are masked out from the query sentence by jointly reasoning over the query sentence and the relevant evidence (Figure~\ref{fig:example}). In addition to text, we also include tables as evidence to further empower LMs to reason over hybrid contexts. 

\nop{Inspired by the QA task, which has become an important testbed for evaluating how well NLP models can understand and reason with languages \cite{xx}, we pre-train the model to answer a question based on given evidence. In this work, we simulate questions by masking entities in a query sentence and treat the masked entities as answers, and pre-train the model to extract the answers from relevant evidence.}

One major challenge in developing \ours\ lies in how to create a large set of query-evidence pairs for pre-training.
Unlike existing self-supervised pre-training methods, examples with complex reasoning cannot be easily harvested from naturally occurring texts. 
Instead, we draw inspiration from  \emph{distant supervision}~\cite{mintz2009distant}, which assumes that ``\textit{any sentence containing a pair of entities that are known to participate in a relation is likely to express that relation},'' and generalize it to our setting of multiple pieces of evidence from text and tables.
Specifically, given a query sentence containing an entity pair, if we mask one of the entities, another sentence or table that contains the same pair of entities can likely be used as evidence to recover the masked entity. Moreover, to encourage deeper reasoning, we collect multiple pieces of evidence that are jointly used to recover the masked entities in the query sentence, allowing us to scatter the masked entities among different \nop{sets}{pieces} of evidence to mimic different types of reasoning. Figure~\ref{fig:example} illustrates several examples using such distant supervision. In Ex.\,1, a model needs to check multiple constraints (i.e., intersection reasoning type) and find ``\textit{the \ul{beach soccer} competition that is established in \ul{1998}}.'' In Ex.\,2, a model needs to find ``\textit{the type of the band that released \ul{Awaken the Guardian}},'' by first inferring the name of the band ``\textit{\ul{Fates Warning}}'' (i.e., bridging reasoning type). 

We first replace the masked entities in a query sentence with the \texttt{[QUESTION]} tokens. The new pre-training objective, span reasoning, \nop{is then to extract}\revise{then extracts} the masked entities from the provided evidence. We augment existing LMs like BERT~\cite{devlin-etal-2019-bert} and RoBERTa~\cite{Liu2019RoBERTaAR} by continuing to train them with the new objective, which leads to \ours, a new LM with better reasoning capabilities.\nop{ We use a transformer based encoder~\cite{devlin-etal-2019-bert} to encode the concatenated query sentence and textual evidence.} Then query sentence and textual evidence are encoded via the LM. When tabular evidence is present, we use the structure-aware transformer TAPAS \cite{herzig-etal-2020-tapas} as the encoder to capture the table structure.

\nop{When finetuning for QA tasks, we append the \texttt{[QUESTION]} token to the question and get contextualized representations with the pre-trained encoder. To encourage the model to conduct deep reasoning, we include multiple pieces of evidence in the input, and scatter the answers among them to mimic different types of reasoning chains {(See Figure \ref{XX})}. To enable our model to handle both structured and unstructured data, we consider both tables and texts as evidence, and adopt the recent structure-aware transformer from TAPAS \cite{XX} to help the model understand table structure.
}

We evaluate \ours\ on the extractive QA task, which is arguably the most representative task requiring reasoning about world knowledge. \nop{When finetuning the models pre-trained by \ours\ for extractive QA, we append the \texttt{[QUESTION]} token to the question and get its contextualized representation from the pre-trained encoder.}  We conduct a comprehensive evaluation using a variety of popular datasets: MRQA \cite{fisch-etal-2019-mrqa}, a \nop{text-only} single-hop QA benchmark including six datasets from different domains; HotpotQA \cite{yang-etal-2018-hotpotqa}, a \nop{text-only} multi-hop QA dataset; NQTables, a subset of the Natural Questions dataset \cite{kwiatkowski-etal-2019-natural} where answers can be found in tables; and HybridQA \cite{chen-etal-2020-hybridqa}, a hybrid multi-hop QA dataset that requires reasoning over both tables and text. Under the few-shot setting, \ours\ substantially outperforms the baselines in almost all datasets, demonstrating that the reasoning ability learned from pre-training can easily transfer to downstream QA tasks and generalize well across domains. Under the full-data setting, \ours\ obtains substantial gains in multi-hop and hybrid QA datasets. Despite its simple model architecture, \ours\ achieves similar or better performance compared with more sophisticated state-of-the-art models for each dataset.
\section{Background}
\label{sec:problem}
\nop{changing to the following flow:}

\noindent \textbf{Language model pre-training}. \nop{Effectiveness of Bert and SpanBert, but they tend to memorize world knowledge.} Existing pre-training objectives such as MLMs \cite{devlin-etal-2019-bert, joshi-etal-2020-spanbert} tend to implicitly memorize the learned knowledge in the parameters of the underlying neural network. In this work, we aim to augment pre-training by encouraging a model to \textit{reason} about (instead of memorizing) world knowledge over the given contexts. 

\noindent \textbf{Extractive question answering}. To measure a model’s reasoning ability about world knowledge, we select extractive QA as a downstream task, which is perhaps one of the most representative tasks for this purpose. Given a question $q$ and provided evidence $E$, an extractive QA model $p_\theta(a|q,E)$ aims to select a contiguous span $a$ from $E$ that answers the question, or output a special token if $E$ is not sufficient to answer the question. 

Our approach, \ours, is inspired by this formulation and extends it to language model pre-training. The challenge in defining such a self-supervised task is in the creation of question-evidence pairs from unlabeled data. Moreover, we aim for a generic approach that works for a wide range of extractive QA settings including single-hop and multi-hop reasoning, hybrid contexts with both unstructured texts and structured tables, as well as few-shot settings. We discuss how to address the challenge and achieve this goal in the next two sections. 

\section{Distant Supervision (DS) for Pre-training}
\label{sec:DS}
We use English Wikipedia as our data source for pre-training. We first extract sentences and tables from Wikipedia pages and then identify salient spans {(such as named entities) from them}\nop{\hs{from the sentences? because you are saying `used as answers...'} that can be used as answers}. We apply the idea of distant supervision and match the sentences and tables to form query-evidence pairs, {which are used to create pre-training examples.}

\subsection{Data Collection}
\noindent\textbf{Text}. We first extract paragraphs from Wikipedia pages and split them into sentences. We consider named entities including both real-world entities (e.g., person, location) and temporal and numeric expressions (e.g., date and quantity) as potential answer entities for pre-training. We first identify real-world entities using existing hyperlinks. Since Wikipedia pages generally do not contain links to themselves, we additionally detect such self-mentions by searching the names and aliases of the topic entity for each page. \revise{Temporal and numeric expressions are identified using an existing NER tool\footnote{\url{https://nlp.johnsnowlabs.com/}}.}

\noindent\textbf{\revise{Tables}}. We extract tables that are labeled as <wikitable> from Wikipedia, and only consider tables with no more than 500 cells. First,  real-world entities are detected using existing hyperlinks. Unlike our method employed for textual sentences, we do not use traditional NER tools here as they are not tailored to work well on tables. Instead, for a cell that does not contain hyperlinks, we match the complete cell value with sentences that are closely related to the table, sourced either from the same page or a page containing a hyperlink pointing to the current page.\nop{I wonder if there is a more clear way to rephrase the following sentence. how to get the matched span in the sentence?} If the matched span in the sentence contains a named entity, we consider the same entity as being linked to the cell as well. Otherwise we consider this cell as a unique entity in the table.\nop{Concrete examples may be needed (can be put in Appendix) to explain this process well.}

Please see Appendix~\ref{sec:data_detail} for details about the tools and resources we use.

\nop{perhaps reordered the writing a bit here.  First say we use existing hyperlinks to aim for high-precision entity linking.  Next, say we don't use traditional NER tools as they are not tailored towards tables, therefore cannot provide the level of precision we're comfortable with.  Then describe what we're doing in addition to hyperlinks to increase table entity recall while keeping the high precision.}

\subsection{Query-Evidence Pairing via DS}
{As described in Section \ref{sec:problem}, a standard QA sample is composed of a question, an answer and evidence. The model infers the relationship between the answer and other entities in the question, and extracts it from the evidence.} In this work, we try to simulate such samples in pre-training. Given a sentence with entities, it can be viewed as a question by masking some entities as answers for prediction. The key issue is then how to find evidence that {contains not only the answer entity, but also the relational information for inference.} Here we borrow the idea of distant supervision \cite{mintz-etal-2009-distant}.\nop{The original distant supervision assumption is that \add{``any sentence that contains a pair of entities that participate in a known knowledge base relation is likely to express that relation in some way''.} We remove the KB relation constraint in the original assumption and generalize it as ``any sentence that contains a pair of related entities is likely to express the relations between the entities in some way''. Given a masked ``question'' sentence, we use other sentences (or tables) that contain the masked ``answer'' entity and another entity in the ``question'' as evidence.}  \nop{c.f.\ the intro of this paper https://arxiv.org/pdf/1704.05958.pdf for the proper way to introduce distant supervision}

Given a sentence as a query, we first extract pairs of entities in it. For each entity pair, we then find other sentences and tables that also contain the same pair as evidence. {Since we do not have the known relation constraint {in the original assumption of distant supervision}, we use the following heuristics to collect evidence that has high quality relational knowledge about the entities and is relevant to the query. First, we only consider entity pairs that contain at least one real-world entity. For textual evidence, the entity pair needs to contain the topic entity of the Wikipedia page, which is more likely to have relations to other entities. For tabular evidence, we consider only entity pairs that are in the same row of the table, but they do not need to contain the topic entity, as in many cases the topic entity is not present in the tables. In both cases, the query and evidence should come from the same page, or the query contains a hyperlink pointing to the evidence page. For tabular evidence, we also allow for the case where the table contains a hyperlink pointing to the query page.}

\begin{table}[t]
    \centering
    \resizebox{\linewidth}{!}{
    \begin{tabular}{lccccc}
    \toprule
        Setting&\# queries&\# sent.&\# tab.&\# ent. pairs\\
    \midrule
        Text-only & 7.6M&8.4M&-&5.5M\\
        Hybrid&3.2M&4.3M&0.9M&6.0M\\
    \bottomrule
    \end{tabular}}\vspace{-0.5em}
    \caption{Statistics about the pre-training data.}\vspace{-0.25em}
    \label{tab:pretrain_data}
\end{table}

\begin{table}[t]
    \centering
    \resizebox{\linewidth}{!}{
    \begin{tabular}{lcccc}
    \toprule
        Setting&All the same&One different&All different\\
    \midrule
        Text-only & 50\%&30\%&20\%\\
        Hybrid&60\%&8\%&32\%\\
    \bottomrule
    \end{tabular}}\vspace{-0.5em}
    \caption{\revise{Analysis of pre-training data quality with 50 examples for each setting. \textit{One different} is when  the relation between the selected entities is different from the relation expressed in the query sentence for of the two pieces of evidence.}}\vspace{-1.25em}
    \label{tab:pretrain_data_quality}
\end{table}
\subsection{Pre-training Data Generation}
\label{sec:data_generation}
Given the query-evidence pairs, a naive way to construct pre-training examples is to sample a single piece of evidence for the query, and mask a shared entity as ``answer'', as in \citet{glass-etal-2020-span}. However, this only simulates simple single-hop questions. In this work, we construct complex pre-training examples that require the model to conduct multi-hop reasoning\nop{ to encourage the model to do deep reasoning}. Here we draw inspiration from how people constructed multi-hop QA datasets. Take HotpotQA~\cite{yang-etal-2018-hotpotqa} as an example. It first collected candidate evidence pairs that contain two paragraphs $(A,B)$, with a hyperlink from $A$ to $B$ so that the topic entity of $B$ is a bridging entity that connects $A$ and $B$. Crowd workers then wrote questions based on each evidence pair. Inspired by this process, we combine multiple pieces of evidence in each pre-training example and predict multiple masked entities simultaneously. The detailed process is described below. Figure~\ref{fig:example} shows two examples. For more examples, please check Appendix~\ref{sec:data_detail}.

We start by sampling up to two entity pairs from the query sentence and one piece of evidence\nop{sample} (sentence or table) for each entity pair. We then mask one entity in each pair as the ``answer'' to predict. {The resulting pre-training examples fall into three categories: (1) Two disjoint entity pairs $\{(a,b), (c,d)\}$ are sampled from the query, and one entity from each pair, e.g., $\{a, c\}$, is masked. This is similar to a combination of two single-hop questions. (2) The two sampled entity pairs $\{(a,b),(b,c)\}$ share a common entity $b$, and $b$ is masked. The model needs to find two sets of entities that respectively satisfy the relationship with $a$ and $c$, and take an intersection (Type II in HotpotQA; see Ex. 1 in Figure~\ref{fig:example}). (3) The two sampled entity pairs $\{(a,b),(b,c)\}$ share a common entity $b$, and $\{b,c\}$ are masked. Here $b$ is the bridging entity that connects $a$ and $c$. The model needs to first identify $b$ and then recover $c$ based on its relationship with $b$ (Type I and Type III in HotpotQA; see Ex. 2 in Figure~\ref{fig:example}).}\nop{Note that the selected entity pairs may contain overlapping entities, so the same entity may be selected as the ``answer'', or both entities in a pair may be masked. This naturally imitates different reasoning types in multi-hop QA, like finding the answer that satisfies multiple constraints {\add{(masking the same entity, see the second example in Figure 1)}}, or finding the answer through a bridging entity {\add{(masking both entities in a pair, see the third example in Figure 1)}}.} We also mask an entity from the query that is not shown in the evidence to simulate unanswerable cases. All sampling is done randomly during pre-training. \nop{just confirm that we don't first prepare the data `before' pre-training, right? Yes} \nop{I think this subsection needs to be significantly expanded. It's where \textit{reasoning} comes from and the design is not very intuitive. For example, it's probably not immediately clear to most audience why you'd need to select two entity pairs. You keep saying ``mimicking different reasoning types'', and this should be the place to elaborate on that.}

\subsection{Data Statistics and Analysis}
\revise{We prepare pre-training data for two settings: (1) one with only textual evidence (text-only) and (2) the other including at least one piece of tabular evidence in each sample (hybrid). Some statistics of the collected data are summarized in Table~\ref{tab:pretrain_data}. For the text-only setting, we extract approximately 7.6M query sentences, each containing 2 entity pairs that are matched with 3 different pieces of textual evidence on average.\nop{how many training instances do we have for each setting? 7.6M * 3?, the unique combinations should be more than that, but we consider going over all 7.6M queries as 1 epoch.} For the hybrid setting, we select approximately 3.2M query sentences, each containing 3.5 entity pairs, matched with 5.8 different pieces of evidence on average.}  \nop{we will have 2 pre-trained models respectively for the 2 settings, right?}

\revise{We also conduct an analysis of the pre-training data quality using 50 randomly sampled examples from each setting. We compare the query sentence and the evidence to see if they are expressing the same relation between the selected entities. Results are summarized in Table~\ref{tab:pretrain_data_quality}. We can see that in both settings, almost 70\% of the examples have the desired characteristic\nop{feature} that the evidence contains useful relational knowledge for recovering missing entities in the query sentence.}

\section{Pre-training}
\subsection{Encoder}
\nop{In this work,  textual and tabular evidence is considered. }For the text-only setting, we use the standard transformer encoder in BERT \cite{devlin-etal-2019-bert}. For settings where the input contains tables, we adopt the transformer variant recently introduced in TAPAS \cite{herzig-etal-2020-tapas}, which uses extra token-type embeddings \nop{like row and column embeddings}{(indicating the row/column position of a token)} to model the table structure. 

\subsection{{Span Reasoning Objective}} 
{Now we describe our \textit{span reasoning objective}, which can advance the reasoning capabilities of a pre-trained model.}

Given a sample collected for pre-training as described in Section \ref{sec:data_generation}, we replace the masked entities $\mathcal{A}=\{a_1,\dots,a_n\}$ {($n$$\leq$$3$)} in the query sentence $q$ with special \texttt{[QUESTION]} tokens. The task then becomes recovering these masked entities from the given evidence $E$ (concatenation of the sampled evidence)\nop{did we mention that $E$ is the concatenation of the two evidences?}. Specifically, we first concatenate $q,E$ and add special tokens to form the input sequence as $[\texttt{[CLS]},q,\texttt{[SEP]},E]$, \nop{in pre-training, we don't have $\texttt{[QUESTION]}$ after $q$, right? yes}and get the contextualized representation $\mathbf{x}$ with the encoder. Since we have multiple entities in $q$ masked with \texttt{[QUESTION]}, for each $a_i$, we use its associated \texttt{[QUESTION]}  representation as a dynamic query vector $\mathbf{x}_{a_i}$ to extract its start and end position $s, e$ of $a_i$ in $E$ (i.e., \textit{question-aware} answer extraction). \nop{check the equations.}

\vspace{-1em}
\begin{equation}\small{
\begin{aligned}
P(s|q, E)&=\frac{\exp \left(\mathbf{x}_{s}^{\top} \mathbf{S} \mathbf{x}_{a_i}\right)}{\sum_{k} \exp \left(\mathbf{x}_{k}^{\top} \mathbf{S} \mathbf{x}_{a_i}\right)} \\
P(e|q, E)&=\frac{\exp \left(\mathbf{x}_{e}^{\top} \mathbf{E} \mathbf{x}_{a_i}\right)}{\sum_{k} \exp \left(\mathbf{x}_{k}^{\top} \mathbf{E} \mathbf{x}_{a_i}\right)}
\end{aligned}}
\label{eqn:selection}
\end{equation}
\vspace{-0.5em}

Here $\mathbf{S}, \mathbf{E}$ are trainable parameters. $\mathbf{x}_{a_i}$ is the representation of special token \texttt{[QUESTION]} corresponding to $a_i$; $\mathbf{x}_{k}$ is the representation of the $k$-th token in $E$. If no answer can be found in the provided evidence, we set $s,e$ to point to the \texttt{[CLS]} token. 

The \textit{span reasoning} loss is then calculated as follows:

\vspace{-1.5em}
\begin{equation}\small{
    L_{SR} = -\sum_{a_i \in \mathcal{A}} \left(\mathrm{log}P\left(s_{a_i}|q,E\right)+ \mathrm{log}P\left(e_{a_i}|q,E\right)\right)
}\end{equation}
\nop{
The loss is then calculated as:
\begin{equation}
    L = - \mathrm{log}P(s|q,E)- \mathrm{log}P(e|q,E)
\end{equation}

At inference time, we concatenate $q,E$ and add special tokens to form the input sequence as $[\texttt{[CLS]},q,\texttt{[QUESTION]},\texttt{[SEP]},E]$, and then score all the start, end locations and rank all spans $(s,e)$ by $g(s,e|q,E)$:

\begin{align}
    f_\mathbf{start} = \mathbf{x}_{i}^{\top} \mathbf{S} \mathbf{x}_{q}&,\ \ \ \ f_\mathbf{end} = \mathbf{x}_{j}^{\top} \mathbf{E} \mathbf{x}_{q} \\
    g(s,e|q,E)&= f_\mathbf{start}(s|q,E)\\
    &+f_\mathbf{end}(e|q,E) \nonumber\\
    &-f_\mathbf{start}(\texttt{[CLS]}|q,E) \nonumber\\
    &-f_\mathbf{end}(\texttt{[CLS]}|q,E) \nonumber
\end{align}
}
\nop{
More specifically, we use the base QA model described in Section \ref{sec:baseqa} to predict the start and end location of the masked entity $a_i$ in $E$. 
\nop{Since we have multiple "answers" masked with \texttt{[QUESTION]} tokens, a fixed query vector $\mathbf{x}_q$ as in Eqn. \ref{eqn:selection} is inapplicable. Instead, for each $a_i$, we use the representation $\mathbf{x}_{a_i}$ of the associated \texttt{[QUESTION]} token to replace $\mathbf{x}_q$ as a dynamic query vector.}
The \textit{span reasoning} loss is then calculated as follows:
\begin{equation}\small{
    L_{SR} = -\sum_{a_i \in \mathcal{A}} \left(\mathrm{log}P\left(s_{a_i}|q,E\right)+ \mathrm{log}P\left(e_{a_i}|q,E\right)\right)
}\end{equation}
}

We name this objective as  \textit{span reasoning}, as it differs from the \textit{span prediction/selection} objectives in existing pre-training work such as SpanBert \cite{joshi-etal-2020-spanbert}, Splinter \cite{ram2021fewshot}, and SSPT \cite{glass-etal-2020-span} in the following ways:\nop{(1) Instead of predicting individual tokens as in the masked language modeling objective, we directly pre-train span selection, which is consistent with the downstream QA model. \hs{should we also compare with other span prediction objectives like the one in SpanBert to better stress what we mean by `grounded' span selection? in other words, what is `ungrounded' like?}} 
(1) Unlike SpanBert and Splinter that use single contiguous paragraph as context, where the models may focus on local cues, we encourage the model to do long-range contextualization by including both query and evidence as input, which can come from different passages, and recovering the masked entities {by grounding them} on the evidence $E$. (2) Unlike SSPT, we improve the model's ability to reason across multiple pieces of evidence by including two disjoint pieces of evidence in a single sample and scattering the answer entities among them to mimic different types of reasoning chains. (3) \nop{Unlike all existing work, w}We mimic the scenario where a span cannot be inferred based on the given contexts, by masking entities in $q$ that do not appear in $E$, in which case the model is trained to select the special \texttt{[CLS]} token.

\subsection{Final Objective}
We also include the \textit{masked language modeling} (MLM) objective in pre-training to leverage other tokens in the input that are not entities. In particular, we randomly mask tokens that are not an entity or token in the header row for tables, and use an MLM objective to recover them. Following the default parameters from BERT, we use a masking probability of 15\%.

The final loss\nop{ for pre-training} is the sum of \textit{span reasoning} loss and \textit{masked language modeling} loss. {Following previous work~\cite{glass-etal-2020-span, herzig-etal-2020-tapas}, we initialize with a pre-trained encoder, and extend the pre-training with our objectives. For the text part, we pre-train two models with BERT-Base (denoted as \ours{\scriptsize{B}}) and RoBERTa-Base (denoted as \ours{\scriptsize{R}}); for the table part, we use TAPAS-Base (denoted as \ours{\scriptsize{T}}).} More implementation details of pre-training are included in Appendix~\ref{sec:pretrain_detail}.

\nop{After reading SSPT, I wonder if we should clarify that ours extends pre-training, rather than training from scratch. See Section 4.2 in \url{https://arxiv.org/pdf/1909.04120.pdf}}

\section{Experiments}
\subsection{Datasets}
\begin{table}[t]
    \centering
    \resizebox{\linewidth}{!}{
    \begin{tabular}{lcccc}
    \toprule
         &MRQA&HotpotQA&NQTables&HybridQA\\
         \midrule
         \# train&86136.5&88881&17112&62686\\
         \# dev&-&1566&1901&3466\\
         \# test&9704&7405&1118&3463\\
         \# evidence&1&10&8.7&34.7\\
         \# tokens*&374.9&89.1&289.6&156.3\\
         has text/table&\cmark/\xmark&\cmark/\xmark&\xmark/\cmark&\cmark/\cmark\\
    \bottomrule
    \end{tabular}}\vspace{-0.5em}
    \caption{Dataset statistics. The statistics for MRQA are averaged over all 6 datasets. \# tokens* is the average number of tokens per evidence.\nop{ Item with * is calculated using the training set.} \nop{stating the obvious, but the tokens row is missing.}}\vspace{-1.25em}
    \label{tab:eval_data}
\end{table}
\begin{table*}[t]
    \centering
    \resizebox{0.93\linewidth}{!}{
    \begin{tabular}{llccccccc}
    \toprule
    Train.\ Size &Model     & SQuAD &TriviaQA&NQ&NewsQA&SearchQA&HotpotQA&Average \\
    \midrule
    \multirow{8}{*}{16}&BERT&  9.9\std{0.6}&   15.4\std{1.3}&  20.5\std{1.5}&  6.5\std{1.2}&   16.8\std{1.2}&  9.6\std{1.6}&   13.1\\
&RoBERTa&       10.3\std{1.1}&  21.0\std{3.1}&  22.5\std{2.1}&  6.7\std{2.0}&   23.4\std{3.5}&  11.2\std{1.0}&  15.9\\
&SpanBERT&      15.7\std{3.6}&  27.4\std{4.1}&  24.3\std{2.1}&  8.1\std{1.4}&   24.1\std{3.2}&  16.3\std{2.0}&  19.3\\
&SSPT&  10.8\std{1.2}&  21.2\std{3.8}&  23.7\std{4.1}&  6.5\std{1.9}&   25.8\std{2.6}&  9.1\std{1.5}&   16.2\\
&Splinter&      16.7\std{5.9}&  23.9\std{3.8}&  25.1\std{2.8}&  11.6\std{1.0}&  23.6\std{4.5}&  15.1\std{3.5}&  19.3\\
&Splinter*&\bf{54.6}& 18.9& 27.4& \bf{20.8}& 26.3& \underline{24.0}&28.7\\
&  \ours\scriptsize{B}&     33.2\std{4.0}&  \underline{37.2}\std{2.6}&  \underline{33.1}\std{2.7}&  11.8\std{2.3}&  \bf{46.1}\std{5.2}&  22.4\std{2.8}&  \underline{30.6}\\
&\ours\scriptsize{R}&  \underline{41.3}\std{5.5}&  \bf{45.5}\std{5.8}&  \bf{33.6}\std{3.9}&  \underline{16.2}\std{3.2}&  \underline{45.8}\std{4.5}&  \bf{34.1}\std{2.9}&  \bf{36.1}\\
\midrule
\multirow{8}{*}{128}&BERT&  21.5\std{1.4}&  23.9\std{0.8}&  31.7\std{0.8}&  11.3\std{1.3}&  32.6\std{2.3}&  14.0\std{0.8}&  22.5\\
&RoBERTa&       48.8\std{4.2}&  36.0\std{2.9}&  36.4\std{2.0}&  22.8\std{2.4}&  41.3\std{2.0}&  35.2\std{1.4}&  36.7\\
&SpanBERT&      61.2\std{4.7}&  48.8\std{6.6}&  38.8\std{2.6}&  31.0\std{5.3}&  50.0\std{3.7}&  44.0\std{2.3}&  45.7\\
&SSPT&  41.5\std{5.0}&  30.3\std{3.7}&  35.0\std{2.4}&  14.0\std{3.6}&  42.8\std{3.5}&  23.7\std{3.4}&  31.2\\
&Splinter&      55.0\std{10.3}& 45.7\std{4.1}&  41.1\std{2.7}&  33.9\std{2.8}&  48.8\std{3.7}&  46.9\std{7.1}&  45.2\\
&Splinter*& \bf{72.7}& 44.7& 46.3& \bf{43.5}& 47.2& \underline{54.7}&\underline{51.5} \\
&  \ours\scriptsize{B}&     58.5\std{2.2}&  \underline{56.2}\std{0.6}&  \underline{46.7}\std{2.6}&  27.8\std{0.6}&  \underline{60.8}\std{1.7}&  45.2\std{2.3}&  49.2\\
&\ours\scriptsize{R}&  \underline{66.7}\std{2.9}&  \bf{62.1}\std{0.9}&  \bf{49.8}\std{1.6}&  \underline{35.7}\std{1.5}&  \bf{62.3}\std{1.7}&  \bf{57.2}\std{0.6}&  \bf{55.6}\\
\midrule
\multirow{8}{*}{1024}&BERT&  64.1\std{0.9}&  41.6\std{2.6}&  50.1\std{0.6}&  43.0\std{0.3}&  53.1\std{1.0}&  46.5\std{1.9}&  49.7\\
&RoBERTa&       77.9\std{0.5}&  62.2\std{1.3}&  60.3\std{0.6}&  55.0\std{0.5}&  67.5\std{0.8}&  63.4\std{0.8}&  64.4\\
&SpanBERT&      \underline{81.1}\std{0.7}&  67.0\std{1.0}&  63.2\std{0.9}&  \underline{56.4}\std{0.4}&  70.0\std{0.8}&  67.6\std{1.1}&  67.5\\
&SSPT&  77.6\std{1.4}&  60.1\std{2.0}&  58.7\std{0.7}&  52.8\std{1.1}&  65.9\std{0.8}&  63.3\std{1.6}&  63.1\\
&Splinter&      79.8\std{3.5}&  67.3\std{1.5}&  63.8\std{0.5}&  54.6\std{1.4}&  68.9\std{0.3}&  68.4\std{1.2}&  67.1\\
&Splinter*& \bf{82.8}& 64.8& \bf{65.5}& \bf{57.3}& 67.3& \bf{70.3}&\underline{68.0}\\
&  \ours\scriptsize{B}&     76.9\std{0.5}&  \underline{67.4}\std{0.5}&  63.6\std{0.6}&  52.2\std{0.5}&  \underline{70.6}\std{0.6}&  67.8\std{0.5}&  66.4\\
&\ours\scriptsize{R}&  79.7\std{0.3}&  \bf{70.1}\std{0.2}&  \underline{65.0}\std{0.9}&  54.7\std{0.6}&  \bf{72.8}\std{0.4}&  \underline{69.7}\std{0.6}&  \bf{68.7}\\
      \midrule
      \multirow{8}{*}{All}&BERT&  88.8&  73.6&  78.7&  67.5&  82.0&  76.2&  77.8\\
&RoBERTa&       92.0&  78.1&  80.6&  \bf{71.9}&  \underline{85.2}&  79.1&  81.2\\
&SpanBERT&      \bf{92.5}&  \bf{79.9}&  80.7&  71.1&  84.8&  \bf{80.7}&  \bf{81.6}\\
&SSPT&  91.1&  77.0&  80.0&  69.7&  83.3&  79.7&  80.1\\
&Splinter&      \underline{92.4}&  \underline{79.7}&  80.3&  70.8&  84.0&  \underline{80.6}&  81.3\\
&Splinter*&92.2& 76.5& \bf{81.0}& 71.3& 83.0& \bf{80.7}&80.8\\
&  \ours\scriptsize{B}&     90.3&  77.5&  79.9&  68.7&  83.7&  80.5&  80.1\\
&\ours\scriptsize{R}&  91.4&  78.9&  \underline{80.8}&  \underline{71.4}&  \bf{85.3}&  \underline{80.6}&  \underline{81.4}\\
      \bottomrule
    \end{tabular}}\vspace{-0.5em}
    \caption{Results on MRQA datasets. \textbf{Best} and \underline{Second Best} results are highlighted. We report the average F1 score over five runs for each dataset, and the macro-average of the six datasets. Splinter* is the result reported in the original paper, where the authors use a deeper model with additional transformation layers on top of the encoder.}\vspace{-1.25em}
    \label{tab:mrqa}
\end{table*}
We conduct \nop{thorough }experiments with a wide range of extractive QA datasets.\nop{To answer the three questions above, we experiment with various datasets for extractive QA.} Statistics are summarized in Table \ref{tab:eval_data}.

\noindent\textbf{MRQA}~\cite{fisch-etal-2019-mrqa}. A single-hop extractive QA benchmark that unifies various existing QA datasets into the same format. Here we use the in-domain subset that contains 6 datasets: SQuAD \cite{rajpurkar-etal-2016-squad}, NewsQA \cite{trischler-etal-2017-newsqa}, TriviaQA \cite{joshi-etal-2017-triviaqa}, SearchQA \cite{Dunn2017SearchQAAN}, HotpotQA \cite{yang-etal-2018-hotpotqa} and Natural Questions \cite{kwiatkowski-etal-2019-natural}. Similar to \citet{ram2021fewshot}, we adapt these datasets to the few-shot setting by randomly sampling smaller subsets from the original training set for training, and use the original development set for testing.
\nop{Do we need to say something about comparing with pre-training methods like REALM somewhere? I don't think it is necessary to compare with them given that they have different focuses.}

\noindent\textbf{HotpotQA}~\cite{yang-etal-2018-hotpotqa}. A multi-hop QA dataset that requires reasoning over multiple pieces of evidence. Here we follow the distractor setting{, where 10 paragraphs are provided to answer a question while only two of them contain relevant information}. We split 10\% of the original train-hard split for development, and use the original development set for testing.

\noindent\textbf{NQTables}~\cite{kwiatkowski-etal-2019-natural}. A subset of the Natural Questions dataset, where at least one answer to the question is present in a table. We extract 19,013 examples from the original training set (307,373 examples) and split them with a 9:1 ratio for training and development. The test set is then created from the original development split (7,830 examples) and contains 1,118 examples. Here we only keep tables from the original Wikipedia article as evidence. Similar subsets are also used in \citet{herzig2021open} and \citet{zayats2021representations}.

\noindent\textbf{HybridQA}~\cite{chen-etal-2020-hybridqa}. A multi-hop QA dataset with hybrid contexts. Each example contains a table and several linked paragraphs.\nop{ where it is necessary to jointly reason over the tabular and textual evidence to answer a question.}

\revise{We adopt the evaluation script from MRQA\footnote{\url{https://github.com/mrqa/MRQA-Shared-Task-2019}}, which evaluates the predicted answer using exact match (EM) and token-level F1 metrics.}
\subsection{Baselines}
\nop{We conduct a comprehensive comparison of \ours\ with existing pre-training methods.}

\noindent\textbf{BERT} \cite{devlin-etal-2019-bert}. A deep transformer model pre-trained with masked languge model (MLM) and next sentence prediction objectives.

\noindent\textbf{RoBERTa} \cite{Liu2019RoBERTaAR}. An optimized version of BERT that is pre-trained with \nop{enlarged}\revise{a larger} text corpus.

\noindent\textbf{SpanBERT} \cite{joshi-etal-2020-spanbert}. A pre-training method designed to better represent and predict spans of text. It extends BERT by masking contiguous random spans, and training the span boundary representation to predict the entire masked span.

\noindent\textbf{SSPT} \cite{glass-etal-2020-span}. A pre-training method designed to improve question answering by training on cloze-like training instances. Unlike \ours, SSPT only masks a single span in the query sentence and predicts it based on an evidence paragraph provided by a separate retriever.

\noindent\textbf{Splinter} \cite{ram2021fewshot}. A pre-training method optimized for few-shot question answering, where the model is pre-trained by masking and predicting recurring spans in a passage.

\noindent\textbf{TAPAS} \cite{herzig-etal-2020-tapas}. A pre-training method designed to learn representations for tables.\nop{TAPAS extends BERT's architecture with extra token-type embeddings (column and row positions) to encode tables.} The model is pre-trained with MLM on tables and surrounding texts extracted from Wikipedia.

For fair comparison, in each task, we use the same model architecture with different pre-trained encoders,\nop{. Our base model for extractive QA} {which} is similar to the one used for {span reasoning} in pre-training. We append the \texttt{[QUESTION]} token to a question and construct the input sequence the same way as in pre-training. We then score all the start, end locations and rank all spans $(s,e)$ (See Eqn.~\ref{eqn:score_0} and \ref{eqn:score_1} in Appendix). {We use a pre-trained encoder and learn the answer extraction layers ($\mathbf{S}, \mathbf{E}$ in Eqn.~\ref{eqn:selection}) from scratch during fine-tuning.} 

Unless otherwise stated, we use the pre-trained base version\nop{`the base pre-trained version' seems not clear; is there another way saying this?} so that all models have similar capacity (110M parameters for \ours{\scriptsize{B}}, 125M parameters for \ours{\scriptsize{R}}, and 111M parameters for \ours{\scriptsize{T}}).  
\subsection{{Few-shot} Single-hop Text QA}
\nop{we might want to justify a bit why we care about few-shot setting or why it is used to demo pre-trained LMs.}
We first experiment with the {easier,} single-hop MRQA benchmark under the few-shot setting to show that our pre-training approach learns general knowledge that can be transferred to downstream QA tasks effectively. Results are shown in Table~\ref{tab:mrqa}. We can see that \ours\ outperforms pre-trained language models such as BERT, RoBERTa and SpanBERT by a large margin on all datasets, particularly with an average absolute gain of 20.3\% and 14.5\% over BERT and RoBERTa respectively. Compared with pre-training methods such as SSPT and Splinter\nop{that are specifically designed for question answering}, \ours\ also shows superior performance and obtains the best results on average\nop{how about splinter*?}. \revise{Under the full-data setting, \ours\ performs competitively and all methods achieve similarly high accuracy. We still demonstrate improvements upon BERT and RoBERTa, and \ours{\scriptsize{R}} second best average score.\nop{Under the full-data setting, \ours\ performs competitively and all methods achieve similarly high accuracy. Please refer to {Table~\ref{tab:mrqa_full} in Appendix} for more details. }\nop{explain a bit how they work and why we are better.}}

\begin{table}[t]
    \centering
    \resizebox{\linewidth}{!}{
    \begin{tabular}{lcccccc}
    \toprule
        \multirow{2}{*}{Model} & \multicolumn{2}{c}{Recall}& \multicolumn{2}{c}{1\%} & \multicolumn{2}{c}{Full}\\
         & Top 2&Top 3&F1&EM&F1&EM\\
         \midrule
         HGN\scriptsize{RoBERTa-Large}&-&-&-&-&82.2&-\\
         HGN\scriptsize{BERT}&-&-&-&-&74.8&-\\
         BERT&92.4&96.9&39.8&28.6&71.9&57.9\\
         RoBERTa&93.1&97.5&56.0&43.1&76.3&62.9\\
         SpanBERT&93.6&97.7&56.5&44.1&76.3&62.9\\
         SSPT&93.9&97.9&54.7&41.8&75.4&61.5\\
         Splinter&94.1&97.9&57.0&44.2&76.5&62.5\\
         \ours\scriptsize{B}&93.8&97.8&57.6&45.3&77.2&63.4\\
         \ours\scriptsize{R}&\bf94.0&\bf98.0&\bf63.1&\bf50.2&\bf78.1&\bf64.8\\
         \bottomrule
    \end{tabular}}\vspace{-0.5em}
    \caption{Results on HotpotQA.}\vspace{-0.5em}
    \label{tab:hotpotqa}
\end{table}
\begin{figure}[t]
    \centering
    \includegraphics[width=0.85\linewidth]{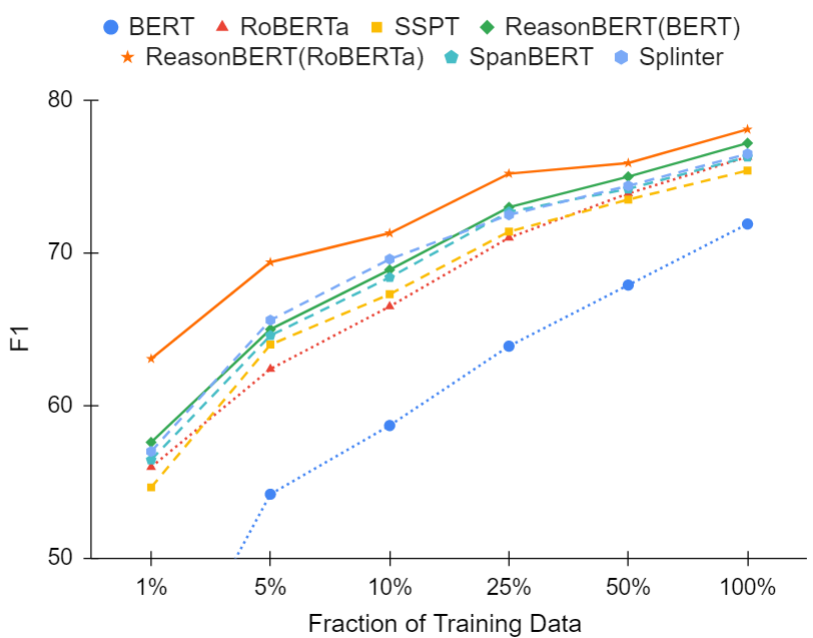}\vspace{-0.5em}
    \caption{Few-shot learning results on HotpotQA.}\vspace{-1em}
    \label{fig:hotpotqa_fewshot}
\end{figure}
\subsection{Multi-hop Text QA}
To demonstrate that our approach is useful in conducting deep reasoning over multiple contexts, we experiment with the HotpotQA dataset. Here we design a simplified multi-hop QA model that first selects relevant paragraphs as evidence, and then extracts the answer from {the top selected evidence}\nop{the selected evidence samples}\nop{do we use `evidence samples' throughout the paper? or `pieces of evidence'}. \revise{Please see Appendix~\ref{sec:finetune_detail} for implementation details.}\nop{Specifically, we first generate all possible paragraphs by sliding a 200-token window over all articles with a stride of 128 tokens. We then train an evidence selector to pick the top 3 evidence samples. As the information for answering a question in HotpotQA is scattered in two articles, we list all possible combinations of paragraphs that come from two different articles and concatenate them together to form the final evidence. We then use the base QA model to extract the answer based on the question and the combined evidence.} In addition to comparing \ours\ with other pre-training methods using the same base model, we also show results for HGN \cite{fang-etal-2020-hierarchical}, which is one of the top ranked models on the HotpotQA leaderboard that uses a more sophisticated model design. 

Results are shown in Table \ref{tab:hotpotqa}. All models perform very well for evidence selection, with over 96\% top 3 recall, but \ours\ still maintains a slim lead over baselines. \ours\ provides a 5.3\% improvement for BERT and a 1.8\% improvement for RoBERTa on overall F1 score, and outperforms all other pre-training methods. \ours\ also outperforms the HGN model with BERT, but is lower than the one using RoBERTa-Large, which is probably due to simpler design and smaller size of the model. We further experiment under the few-shot setting. Here we focus on the QA performance, so we reuse the evidence selector trained with full data for each model, and train the QA module with different fractions of training data. We can see that the advantage of using \ours\ is more obvious with limited training data. With 1\% of training data, \ours{\scriptsize{R}} obtains F1 score of 63.1\%, a 7.1\% absolute gain over RoBERTa. \revise{Results for training the QA model with different fraction of training data is shown in Figure~\ref{fig:hotpotqa_fewshot}. We can see that \ours\ obtains larger gain under the few-shot setting.}\nop{ Please see {Appendix~\ref{sec:finetune_detail} and \ref{sec:hotpotqa_fewshot}} for implementation details and the full few-shot results.}
\nop{In Section 6.3-6.5, can we move some details to appendix and leave enough space to discuss results.}

\subsection{Table QA}
\begin{figure}[t]
    \centering
    \includegraphics[width=0.85\linewidth]{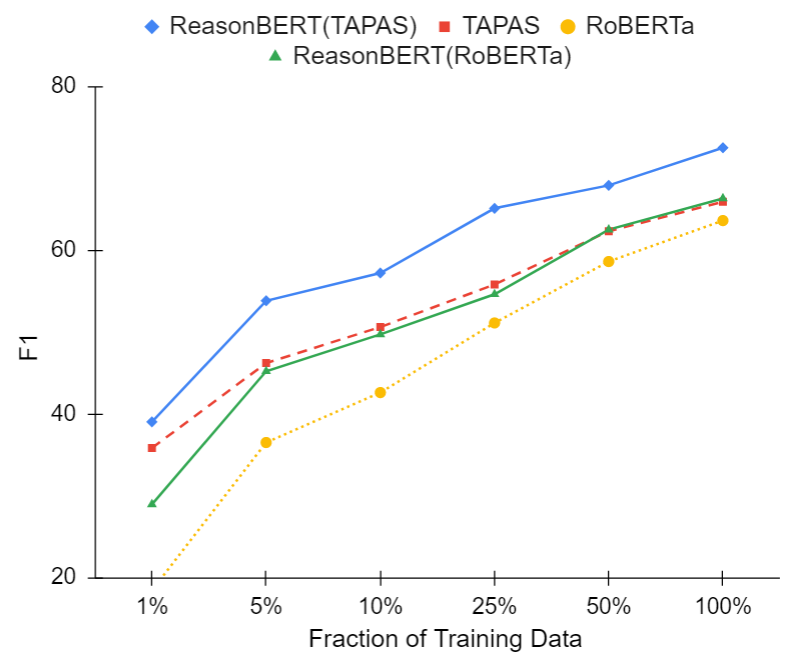}\vspace{-0.5em}
    \caption{Few-shot learning results on NQTables.}\vspace{-0.5em}
    \label{fig:nqtables_fewshot}
\end{figure}

We demonstrate our approach also works with structured data such as tables using the NQTables dataset. We first use a text based RoBERTa encoder as baseline, which linearizes a table as a text sequence, by concatenating tokens row by row and separating cells with the \texttt{[SEP]} token.\nop{We split the text sequence to fit the max input length by sliding a window with a stride of 128 tokens.} We then experiment with the structure-aware encoder from TAPAS and compare the pre-trained TAPAS encoder with the one pre-trained using \ours.\nop{ Here we truncate each cell to 50 tokens, and split the table into snippets horizontally. Same as pre-training, we include the first row and column in each table snippet.} Results are shown in Table \ref{tab:nqtable}. First, we can see that TAPAS outperforms RoBERTa by 2.3\%, demonstrating the importance of modeling the table structure. \ours{\scriptsize{R}} slightly outperforms TAPAS {on test set}, but \ours{\scriptsize{T}} further boosts F1 to 72.5\%, resulting in {at least} 6.6\% absolute gains over existing methods. \revise{Results for training the Table QA model with different fractions of training data are shown in Figure~\ref{fig:nqtables_fewshot}. \ours{\scriptsize{T}} consistently outperforms TAPAS while \ours{\scriptsize{R}} gradually matches the performance of TAPAS with the increasing of training data.}\nop{Given the description here, should we adjust the order of models in Table 5? Put ours as the last two lines?}
\begin{table}[t]
    \centering
    \resizebox{0.8\linewidth}{!}{
    \begin{tabular}{lcccc}
    \toprule
        \multirow{2}{*}{Model} & \multicolumn{2}{c}{Dev} &\multicolumn{2}{c}{Test}\\
         & F1&EM&F1&EM\\
         \midrule
         RoBERTa&58.9&52.8&63.6&58.1\\
         \ours\scriptsize{R}&61.9&56.4&66.3&60.9\\
         TAPAS&64.9&57.8&65.9&59.6\\
         \ours\scriptsize{T}&\bf69.2&\bf63.5&\bf72.5&\bf67.3\\
         \bottomrule
    \end{tabular}}\vspace{-0.5em}
    \caption{Results on NQTables.}\vspace{-1.25em}
    \label{tab:nqtable}
\end{table}
\subsection{Hybrid QA}
We further evaluate our approach on HybridQA, a multi-hop question answering dataset using both text and tables as evidence. \citet{chen-etal-2020-hybridqa} proposes a baseline model HYBRIDER that divides the problem into four tasks: linking, ranking, hopping and reading comprehension.\nop{ 1) linking: link questions to their corresponding cells use heuristics. 2) ranking: rank the linked cells use a neural model. 3) hopping: based on the cell selected in the last step, decide which neighboring cell or itself contains the final answer. 4) reading comprehension: extract the answer from the predicted cell or its linked paragraph.} We follow their design but simplify the model by merging ranking and hopping into a single cell selection task. We use the linking results from \citet{chen-etal-2020-hybridqa},\nop{ For each linked cell, we take a snippet out of the original table including the headers, the entire row of the linked cell, and concatenate the evidence sentence to the cell if it is linked through the hyperlinked passage.} and then train a table based cell selector to select the cell which is the answer or is linked to the passage that contains the answer. Finally, we train a text based QA model to extract the final answer by taking the table snippet that contains the selected cell, and concatenating it with the hyperlinked passage as evidence. \revise{Please see Appendix~\ref{sec:finetune_detail} for implementation details.}
Results are shown in Table \ref{tab:hybridqa}. First, we can see that our simplified architecture works surprisingly well, with TAPAS for cell selection and RoBERTa for QA, we already outperform HYBRIDER. The performance is further improved by replacing the encoders with \ours{\scriptsize{T}} and \ours{\scriptsize{R}}, and substantially outperforms the best model on the leaderboard (52.04 EM) at the time of submission.\nop{should we put this result in table 6?}
\begin{table}[t]
    \centering
    \resizebox{\linewidth}{!}{
    \begin{tabular}{lcccccc}
    \toprule
        \multirow{2}{*}{Model} & \multicolumn{2}{c}{Cell Selection}&\multicolumn{2}{c}{Dev}&\multicolumn{2}{c}{Test}\\
        &Top 1&Top 2&F1&EM&F1&EM\\
         \midrule
         HYBRIDER\scriptsize{BERT-Base}&-&-&50.9&43.7&50.2&42.5\\
         HYBRIDER\scriptsize{BERT-Large}&68.5&-&50.7&44.0&50.6&43.8\\
         TAPAS+RoBERTa&73.3&79.7&64.0&57.3&63.3&56.1\\
         \ours&\bf76.1&\bf81.3&\bf67.2&\bf60.3&\bf65.3&\bf58.0\\
         \bottomrule
    \end{tabular}}\vspace{-0.5em}
    \caption{Results on HybridQA.}\vspace{-1em}
    \label{tab:hybridqa}
\end{table}
\begin{table}[t]
    \centering
    \resizebox{\linewidth}{!}{
    \begin{tabular}{lcccc}
    \toprule
        \multirow{2}{*}{Model}  &\multicolumn{2}{c}{1024}&\multicolumn{2}{c}{Full}\\
         &F1&EM&F1&EM\\
         \midrule
         \ours{\scriptsize{R}}&65.2&52.8&79.2&65.8\\
         \hspace{2em}-- MLM&63.7&51.3&77.7&64.0\\
         \hspace{2em}-- Unanswerable Ent.&64.4&51.8&78.4&65.0\\
         \hspace{2em}-- Multiple Evidences&60.8&48.6&77.8&64.5\\
         \bottomrule
    \end{tabular}}\vspace{-0.5em}
    \caption{Ablation study on HotpotQA.}\vspace{-1.25em}
    \label{tab:ablation}
\end{table}
\section{Ablation Study}
We further conduct ablation studies on HotpotQA to verify our design choices, summarized in Table \ref{tab:ablation}. Here we remove different components of \ours{\scriptsize{R}} and test them under both the full-data and few-shot setting (with 1024 examples). To save computing resources, here all models are pre-trained with 5 epochs. We can see that combining multiple pieces of evidence and predicting multiple masked spans\nop{can we see this multiple masked spans thing?} simultaneously brings the most gain, especially under the few-shot setting. This  is probably because the setting allows us to simulate complex reasoning chains and encourage the model to do deep reasoning. Masking unanswerable entities and utilizing MLM also help to improve performance.
\section{Related Work}
\noindent\textbf{Language Model Pre-training.} Contextualized word representations pre-trained on large-scale unlabeled text corpus have been widely used in NLP lately. Most prevalent approaches are variants of pre-trained language models such as BERT~\cite{devlin-etal-2019-bert} and RoBERTa~\cite{Liu2019RoBERTaAR}.\nop{ BERT first introduces the masked language model pre-training objective, which is inspired by the Cloze task, and shows promising results for applying pre-training/fine-tuning paradigm on NLP tasks. RoBERTa further improves upon BERT by enlarging the pre-training corpus and extending the pre-training steps.}
\revise{More recently, self-supervised pre-training has also shown promising results on modalities other than plain text, such as tables~\cite{herzig-etal-2020-tapas,deng2020turl,iida-etal-2021-tabbie}, knowledge bases~\cite{zhang-etal-2019-ernie, peters-etal-2019-knowledge} and image-text~\cite{su2019vl}.}
\nop{Meanwhile, generative language models such as BART~\cite{lewis-etal-2020-bart} and GPT-3~\cite{NEURIPSgpt3} have also achieved great success in both generation and comprehension tasks. Though effective in learning general language representations, these language models tend to implicitly store world knowledge in the network parameters. As a result, one must train ever-large models to capture more knowledge that can cover different tasks, which can be prohibitively expensive to train and slow to use.}Meanwhile, there has also been work that uses pre-training to accommodate specific needs of downstream NLP tasks, such as open-domain retrieval~\cite{guu2020realm}, representing and predicting spans of text~\cite{joshi-etal-2020-spanbert} and semantic parsing~\cite{yu2020grappa,deng-etal-2021-structure}.\nop{REALM augments language model pre-training with a learned knowledge retriever to improve interpretability and modularity, and outperforms all previous systems in Open-domain Question Answering by a large margin. SpanBERT aims to better represent and predict spans of text. Instead of masking individual tokens, SpanBERT masks contiguous random spans, and train the span boundary representations to predict the entire content of the masked span. The experimental results show that SpanBERT brings substantial improvement on tasks like question answering and co-reference resolution.}

\noindent\textbf{Machine Reading Comprehension.} Machine reading comprehension (MRC) {or extractive QA} has become an important testbed for natural language understanding evaluation~\cite{fisch-etal-2019-mrqa}.\nop{ In a typical MRC setting, the system needs to answer a question based on one or more provided pieces of evidence, which requires the system to explicitly reason over multiple contexts.} The conventional method to train an MRC model usually relies on large-scale supervised training data~\cite{chen-etal-2017-reading, Zhang2020RetrospectiveRF}. Recently, more and more work has focused on developing self-supervised methods that can reduce the need for labeled data for more efficient domain adaptation, while achieving the same or even better performance. One direction is question generation~\cite{pan-etal-2021-MQA-QG}, which automatically generates questions and answers from unstructured and structured data sources using rules or neural generators. Recent work also tries to directly simulate questions with cloze-like query sentences. Splinter~\cite{ram2021fewshot} proposes to pre-train the model by masking and predicting recurring spans. However, this limits the query and context to come from the same passage. In contrast, SSPT~\cite{glass-etal-2020-span} also pre-trains with a span selection objective, but uses a separate document retriever to get relevant paragraphs as context.

Our work is most related to SSPT, but uses distant supervision to collect query-evidence pairs and thus obviate the need for a retriever. Meanwhile, to encourage the model to learn complex reasoning, we mimic different types of reasoning chains by masking multiple entities, including unanswerable ones, and simultaneously inferring them from disjoint pieces of evidence. Our method also works with heterogeneous sources including both text and tables, while most existing work considers only text-based question answering.

\section{Conclusion and Future Work}
We propose \ours, a novel pre-training method to enhance the reasoning ability of language models. The resulting model obtains substantial improvements on multi-hop and hybrid QA tasks that require complex reasoning, and demonstrates superior few-shot performance. In the future, we plan to use our query-evidence pairs collected by distant supervision to improve the retrieval performance for open-domain QA, as well as empower \ours\ to handle more types of reasoning, like comparison and numeric reasoning, in natural language understanding. 

\section*{Acknowledgements}
We would like to thank the anonymous reviewers for their helpful comments. Authors at The Ohio State University were sponsored in part by Google Faculty Award, the Army Research Office under cooperative agreements W911NF-17-1-0412, NSF Grant IIS1815674, NSF CAREER \#1942980, Fujitsu gift grant, and Ohio Supercomputer Center \cite{OhioSupercomputerCenter1987}. The views and conclusions contained herein are those of the authors and should not be interpreted as representing the official policies, either expressed or implied, of the Army Research Office or the U.S. Government. The U.S. Government is authorized to reproduce and distribute reprints for Government purposes notwithstanding any copyright notice herein. Research was also supported with Cloud TPUs from Google's TPU Research Cloud (TRC).
\bibliography{anthology,custom}
\bibliographystyle{acl_natbib}

\clearpage
\appendix

\section{Implementation Details}
\subsection{Pre-training Data Details}
\label{sec:data_detail}
We extract paragraphs from Wikipedia XML dump\footnote{\url{https://dumps.wikimedia.org/}} use JWPL\footnote{\url{https://dkpro.github.io/dkpro-jwpl/}} and tables use wikitextparser\footnote{\url{https://github.com/5j9/wikitextparser}}. The paragraphs are then processed with SparkNLP\footnote{\url{https://nlp.johnsnowlabs.com/}} for sentence boundary detection and named entity recognition. 

Table~\ref{tab:data_example_text} and Table~\ref{tab:data_example_hybrid} show some examples of the query-evidence pairs we collected for pre-training. The selected entities are underlined. During pre-training, we will mask some of the entities in the query and recover them based on the evidence. As the pre-training data is collected via distant supervision, it contains some noise. Here we also include some bad examples where the evidence does not express the same relation between the selected entities as the query sentence (highlighted in red).

\subsection{Pre-training Details}
\label{sec:pretrain_detail}
We set the max length of query sentences to 100 tokens and the max length of \nop{each evidence sample}\revise{single piece of evidence} to 200 if there are two evidence selections or 400 if there is only one. For textual evidence, we include the neighbouring sentences from the same paragraph as extra context for the selected evidence sentence and clip to the max evidence length. For tabular evidence, we take a snippet of the original table, and truncate the cells to 20 tokens. We always keep the first row and column in the table, as they often contain important information such as headers and subject entities. Based on the selected entity pair, we sample up to 5 columns and include as many rows as possible until reaching the budget.

We initialize {our encoder} with BERT-Base\footnote{\url{https://huggingface.co/bert-base-uncased/tree/main}} and RoBERTa-Base\footnote{\url{https://huggingface.co/roberta-base/tree/main}} for the text part\nop{-only setting}, and TAPAS-base\footnote{\url{https://huggingface.co/google/tapas-base/tree/no_reset}} for the table part\nop{hybrid setting}. We train \ours\ using AdamW \cite{Loshchilov2019DecoupledWD} for 10 epochs with batches of 256 sequences of length 512; this is approximately 290k steps with text-only data, and 120k steps with hybrid data. We base our implementation on Huggingface Transformers \cite{wolf-etal-2020-transformers}, and train on a single eight-core TPU on the Google Cloud Platform.

\subsection{Fine-tuning Details}
\label{sec:finetune_detail}
To extract the answer span from given evidence, we score all the start, end locations and rank all spans $(s,e)$ by $g(s,e|q,E)$ as follows:

{\small{\begin{align}
    f_\mathbf{start} = \mathbf{x}_{s}^{\top} \mathbf{S} \mathbf{x}_{q}&,\ \ \ \ f_\mathbf{end} = \mathbf{x}_{e}^{\top} \mathbf{E} \mathbf{x}_{q} \label{eqn:score_0}\\
    g(s,e|q,E)&= f_\mathbf{start}(s|q,E)\label{eqn:score_1}\\
    &+f_\mathbf{end}(e|q,E) \nonumber\\
    &-f_\mathbf{start}(\texttt{[CLS]}|q,E) \nonumber\\
    &-f_\mathbf{end}(\texttt{[CLS]}|q,E) \nonumber
\end{align}}}

For all fine-tuning experiments, we set the batch size to 20 and use a maximal learning rate of $5 \cdot 10^{-5}$, which warms up in the first 10\% of the steps, and then decays linearly. We use the development set for model selection if it is present, otherwise we use the last model checkpoint. 

\noindent\textbf{Single-hop text QA.}
We split the text sequence to fit the max input length by sliding a window with a stride of 128 tokens.

For the few-shot setting, we fine-tune the model for either 10 epochs or 200 steps (whichever is larger). For the fully supervised setting, we fine-tune the model for 2 epochs.

\noindent\textbf{Multi-hop text QA.}
We design a simplified multi-hop QA model that first selects relevant paragraphs as evidence, and then extracts the answer from the selected evidence samples. Specifically, we first generate all possible paragraphs by sliding a 200-token window over all articles with a stride of 128 tokens. We then train an evidence selector to pick the top 3 evidence samples. As the information for answering a question in HotpotQA is scattered in two articles, we list all possible combinations of paragraphs that come from two different articles and concatenate them together to form the final evidence. We then use the base QA model to extract the answer based on the question and the combined evidence.

We fine-tune the evidence selector model for 2 epochs, and the QA model for 5 epochs with full data. For the few-shot setting, we fine-tune the QA model for 10 epochs with 1\&, 5\% and 10\% of the training data, and for 5 epochs with 25\% and 50\% of the training data.

\noindent\textbf{Table QA.}
For the text based model, We split the text sequence to fit the max input length by sliding a window with a stride of 128 tokens. For the table based model, we truncate each cell to 50 tokens, and split the table into snippets horizontally. Same as pre-training, we include the first row and column in each table snippet.

We fine-tune the model for 5 epochs with full data. For the few-shot setting, we fine-tune the QA model for 10 epochs with 1\&, 5\% and 10\% of the training data, and for 5 epochs with 25\% and 50\% of the training data.

\noindent\textbf{Hybrid QA.}
\citet{chen-etal-2020-hybridqa} proposes a baseline model that divides the problem into four tasks: 1) linking: link questions to their corresponding cells using heuristics. 2) ranking: rank the linked cells use a neural model. 3) hopping: based on the cell selected in the last step, decide which neighboring cell or itself contains the final answer. 4) reading comprehension: extract the answer from the predicted cell or its linked paragraph. We follow their design and simplify the model by merging ranking and hopping into a single cell selection task. We use the linking results from \citet{chen-etal-2020-hybridqa}. For each linked cell, we take a snippet out of the original table including the headers, the entire row of the linked cell, and concatenate the evidence sentence to the cell if it is linked through the hyperlinked passage. To select the cell, we train the model to select separately on the token, row and column level, and aggregate the final scores . More specifically, we calculate the probability of selecting on the token and row level as follows:
\begin{equation}\small{
\begin{aligned}
    P(t|q, E)&=\frac{\exp \left(\mathbf{x}_{t}^{\top} \mathbf{S} \mathbf{x}_{a_i}\right)}{\sum_{k} \exp \left(\mathbf{x}_{k}^{\top} \mathbf{S} \mathbf{x}_{a_i}\right)} \\
    S_{cell}&=\mathrm{mean}_{x_i \in cell}\left(\mathbf{x}_{i}^{\top} \mathbf{R} \mathbf{x}_{a}\right)\\
    P(r_a=j \mid q, E)&=\frac{\exp \left(\max_{cell \in r_j}S_{cell}\right)}{\sum_{k} \exp \left(\max_{cell \in r_k}S_{cell}\right)}
\end{aligned}
}\end{equation}
Here $\mathbf{S}$ is the weight matrix of the token selection header, we only consider the first token in each cell, and $t$ is the first token of the selected cell. $\mathbf{R}$ is the weight matrix of row selection header, and the column selection probability is calculated similarly with another column selection header. We first score each cell by averaging over all tokens in that cell. We then do a max pooling over all cells in the row or column so the model can focus on the strongest signal, for example the column header. The final probability of selecting a cell is the sum of token, row and column scores.

The input for the QA model then contains the header of the table, the row of the selected cell, and the hyperlinked passage.

We fine-tune the cell selection model for 2 epochs and the QA model for 3 epochs.

\begin{table*}[t]
    \centering
    \resizebox{\linewidth}{!}{
    \begin{tabular}{
       >{\raggedright}p{6cm}
       >{}p{\dimexpr\textwidth-6\tabcolsep-4\fboxsep-4.5cm\relax}
    }
    \toprule
    Query
      & Evidence \\
    \midrule
    \multirow{2}{6cm}{
    "\underline{\it I Thought I Lost You}" was nominated for Broadcast Film Critics Association Award for Best Song and Golden Globe Award for Best Original Song, but lost both to \underline{\it Bruce Springsteen}'s "\underline{\it The Wrestler}" from  \underline{\it The Wrestler} (2008).
    } &
    On January 11, 2009, \underline{\it Springsteen}  won the Golden Globe Award for Best Song for "\underline{\it The Wrestler}", from the Darren Aronofsky \underline{\it film by the same name}.
\\
      &"\underline{\it I Thought I Lost You}" was nominated for the Broadcast Film Critics 
Association Award for Best Song at the 14th Broadcast Film Critics Association Award, but 
lost to \underline{\it Bruce Springsteen}'s "\underline{\it The Wrestler}" from  \underline{\it The Wrestler} (2008).\\
    \midrule
    
    \multirow{2}{6cm}{
    Film critic \underline{\it Roger Ebert}  compared it to \underline{\it John Carpenter}'s  \underline{\it Halloween}, noting: "Blue Steel" is a sophisticated update of Halloween, the movie that first made  \underline{\it Jamie Lee Curtis} a star.
    } &
    Historian Nicholas Rogers notes that film critics contend that Carpenter's direction and 
camera work made \underline{\it Halloween} a "resounding success." \underline{\it Roger Ebert} remarks, ...
\\
      &Since \underline{\it Jamie Lee Curtis}, the main actress 
from the original and the sequel Halloween II (1981), wanted to reunite the cast and crew of 
the original film, she asked  \underline{\it Carpenter} to direct Halloween H20: 20 Years Later.\\
    \midrule
    
    \multirow{2}{6cm}{
    A hybrid disc is an optical disc that has multiple \underline{\it file system} installed on it, 
typically \underline{\it ISO 9660} and HFS+ (or  \underline{\it HFS} on older discs).
    } &
    Hierarchical File System  ( \underline{\it HFS} ) is a proprietary  \underline{\it file system}  developed by Apple Inc. for
use in computer systems running Mac OS.
\\
      &\underline{\it ISO 9660}  is a  \underline{\it file system}  for optical disc media. Being sold by the International Organization for Standardization (ISO) the  file system  is considered an international technical standard.\\
    \midrule
    
    \multirow{2}{6cm}{
    After \underline{\it 1709} , the heads of the \underline{\it House of Orleans} branch of \underline{\it the  House of Bourbon} ranked as the prince of the Blood  – this meant that the dukes could be addressed as Monsieur le Prince (a style they did not, however, use).
    } &
    From  \underline{\it 1709}  until the French Revolution, the  \underline{\it Orleans} dukes were next in the order of succession to the  French  throne after members of the senior branch of the House of Bourbon, descended from Louis XIV.
\\
      &Restored briefly in 1814 and definitively in 1815 after the fall of the First French Empire, the senior line of the Bourbons was finally overthrown in the July Revolution of 1830. A cadet  \underline{\it Bourbon}  branch, the \underline{\it House of Orleans}, then ruled for 18 years (1830–1848), until it too was overthrown.
\\
    \midrule
    
    \multirow{2}{6cm}{
    The \underline{\it Citroen C6}  is an  \underline{\it executive car}  produced by the French car maker  \underline{\it Citroen}  from 2005 to \underline{\it 2012}.
    } &
    The  \underline{\it C6}  was aimed as a stylish alternative to  \underline{\it executive cars}, like the BMW 5 Series and 
the Audi A6, and it has been described as  "spaceship that rides on air", "charmingly 
idiosyncratic" and "refreshingly different".
\\
      &In  \underline{\it\color{red} 2012}, \underline{\it\color{red} Citroen}  announced plans to enter the World Touring Car Championship. The team 
transformed a DS3 WRC into a laboratory vehicle to help with early development, while ...
\\
    \midrule
    
    \multirow{2}{6cm}{
    Leaving the \underline{\it Market Street subway} at Ferry Portal heading south, the T Third Street follows \underline{\it The Embarcadero} south of Market Street, then veers onto King Street in front of \underline{\it Oracle Park} until it reaches the \underline{\it Caltrain}  station terminal.
    } &
    the 4th \& King  \underline{\it\color{red} Caltrain}  station is 1.5 blocks from the 
stadium, and the  \underline{\it\color{red} Oracle Park} Ferry Terminal is outside the east edge of the ballpark beyond the center field bleachers.
\\
      &the southwestern end of the  \underline{\it\color{red} Market Street subway}  connects to the much-older Twin Peaks Tunnel, and the northeastern end  connects to surface tracks along the  \underline{\it\color{red} The Embarcadero}.
\\
    \bottomrule
    \end{tabular}}
    \caption{Pre-training data examples, text-only setting.}
    \label{tab:data_example_text}
\end{table*}

\begin{table*}[t]
    \centering
    \resizebox{\linewidth}{!}{
    \begin{tabular}{
       >{\raggedright}p{6cm}
       >{}p{\dimexpr\textwidth-6\tabcolsep-4\fboxsep-4.5cm\relax}
    }
    \toprule
    Query
      & Evidence \\
    \midrule
    \multirow{2}{6cm}{
    \underline{\it Rowland Barran} (7 August 1858 – 6 August 1949) was an English \underline{\it Liberal} \underline{\it Party}  politician and \underline{\it Member of} \underline{\it Parliament}.} &
    \underline{\it Rowland Barran}  was the youngest son of Sir John Barran, a pioneer in clothing manufacture and \underline{\it Member of Parliament} for Leeds and Otley.
\\
      &\vspace{-15pt}\begin{tabular}[t]{
         >{\raggedright}p{1cm}
         >{\raggedright}p{3.5cm}
         >{}p{4cm}
      }
      \toprule
      Year&Member&Party\\
      \midrule
      1885&William Jackson&Conservative\\
      1902&\underline{\it Rowland Barran}&\underline{\it Liberal}\\
      1918&Alexander Farquharson&Coalition Liberal\\
      \bottomrule
      \end{tabular}\vspace{3pt}\\
    \midrule
    \multirow{2}{6cm}{
    "\underline{\it It Ain't Over 'til It's Over}" is a song recorded, written, and produced by American musician \underline{\it Lenny Kravitz} for his second studio album, Mama Said (1991).} &
    Retrieved on August 19, 2007. \underline{\it Kravitz}'s biggest single yet, "\underline{\it It Ain't Over 'til It's Over}", went to number 2 on the Billboard Hot 100.
\\
      &\vspace{-15pt}\begin{tabular}[t]{
         >{\raggedright}p{2cm}
         >{\raggedright}p{1cm}
         >{\raggedright}p{2.5cm}
         >{}p{2.5cm}
      }
      \toprule
      Act&Order&Song&Rock Artist\\
      \midrule
      In Stereo & 4 & Demons & Imagine Dragons\\
      Cyrus Villanueva & 5 & \underline{\it It Ain't Over} \underline{\it 'til It's Over} & \underline{\it Lenny Kravitz} \\
      Michaela Baranov & 6 & Wild Horses & The Rolling Stones\\
      \bottomrule
      \end{tabular}\vspace{3pt}\\
    \midrule
    \multirow{1}{6cm}{
    \underline{\it Ronnie Bremer} raced the first five races of the season  with \underline{\it Brooks Associates Racing}, before moving to Polestar Motor Racing.}
&\vspace{-15pt}\begin{tabular}[t]{
         >{\raggedright}p{1cm}
         >{\raggedright}p{2.5cm}
         >{}p{5cm}
      }
      \toprule
      Place&Name&Team\\
      \midrule
      5&\underline{\it Ronnie Bremer}&\underline{\it Brooks Associates Racing}\\
      6&Bryan Sellers&Lynx Racing\\
      9&Jonathan Bomarito&Transnet Racing\\
      \bottomrule
      \end{tabular}\vspace{3pt}\\
    \midrule
    \multirow{1}{6cm}{
    Donovan also appeared in the \underline{\it 1980} film \underline{\it Breaker Morant}, but in a subsidiary role, rather than as the title character.}
&\vspace{-15pt}\begin{tabular}[t]{
         >{\raggedright}p{3.5cm}
         >{\raggedright}p{1cm}
         >{}p{4cm}
      }
      \toprule
      Title&Year&Role\\
      \midrule
      Cop Shop (TV series)&1978-1980&Detective Sgt. Vic\\
      \underline{\it Breaker Morant}&\underline{\it 1980}&Captain Simon Hunt\\
      Smash Palace&1981&Traffic Officer\\
      \bottomrule
      \end{tabular}\vspace{3pt}\\
    \midrule
    \multirow{2}{6cm}{
    \underline{\it Try a Little Kindness} is the sixteenth album by American singer/guitarist \underline{\it Glen Campbell}, released in \underline{\it 1970}.} &
    At the height of his popularity, a \underline{\it\color{red} 1970} biography by Freda Kramer, The \underline{\it\color{red} Glen Campbell} Story, was published.
\\
      &\vspace{-15pt}\begin{tabular}[t]{
         >{\raggedright}p{1cm}
         >{\raggedright}p{2.5cm}
         >{\raggedright}p{2.2cm}
         >{}p{2.3cm}
      }
      \toprule
      Day&Album&Artist&Notes\\
      \midrule
      1& On the Boards&Taste&-\\
      26&Chicago&Chicago&aka Chicago II\\
      -&\underline{\it Try a Little} \underline{\it Kindness}&\underline{\it Glen Campbell}&-\\
      \bottomrule
      \end{tabular}\vspace{3pt}\\
    \bottomrule
    \end{tabular}}
    \caption{Pre-training data examples, hybrid setting.}
    \label{tab:data_example_hybrid}
\end{table*}

\end{document}